# Multicolumn Networks for Face Recognition

Weidi Xie  
weidi@robots.ox.ac.uk  
Andrew Zisserman  
az@robots.ox.ac.uk  

Visual Geometry Group  
Department of Engineering Science  
University of Oxford  
Oxford, UK

## Abstract

The objective of this work is set-based face recognition, i.e. to decide if two sets of images of a face are of the same person or not. Conventionally, the set-wise feature descriptor is computed as an average of the descriptors from individual face images within the set. In this paper, we design a neural network architecture that learns to aggregate based on both "visual" quality (resolution, illumination), and "content" quality (relative importance for discriminative classification).

To this end, we propose a Multicolumn Network (MN) that takes a set of images (the number in the set can vary) as input, and learns to compute a fix-sized feature descriptor for the entire set. To encourage high-quality representations, each individual input image is first weighted by its "visual" quality, determined by a self-quality assessment module, and followed by a dynamic recalibration based on "content" qualities relative to the other images within the set. Both of these qualities are learnt implicitly during training for set-wise classification. Comparing with the previous state-of-the-art architectures trained with the same dataset (VGGFace2), our Multicolumn Networks show an improvement of between 2-6% on the IARPA IJB face recognition benchmarks, and exceed the state of the art for all methods on these benchmarks.

## 1 Introduction

Set-based recognition is commonly used in applications where the task is to determine if the two sets belong to the same person or not, for instance, for face verification or person re-identification. In the paper we will focus on the example of human faces, where each set could contain face images of the same person from multiple sources (e.g. still images, video from survillience cameras, or a mixture of both). With the great success of deep learning for image classification [6, 7, 9, 14], the popular approach to set-based face verification is to first train a deep convolutional neural network (CNN) on face image classification, and a set-wise descriptor is then obtained by simply averaging over the CNN feature vectors of individual images in the set [3, 11, 13, 16]. Verification then proceeds by comparing the set-based vectors. With the help of a large, diverse dataset, this approach has already achieved very impressive results on challenging benchmarks, such as the IARPA IJB-A, IJB-B and IJB-C face recognition datasets [8, 10, 18].

The simple averaging combination rule suffers from two principal deficiencies:
*First, visual quality assessment*–the naïve average pooling ignores the difference in the amount of information provided by each face image within the set. For instance, an aberrant image (blurry, extreme illumination or low resolution) will not contribute as much information as an 'ideal' image (sharp focus, well lit and high resolution). Thus, as well as producing a face descriptor for a face image, a "self-aware" quality measure should also be produced reflecting the visual quality of each input image. This quality measure should be taken into account when computing the set-based descriptor.

*Second, content-aware quality assessment*–while computing the set-based feature descriptor, the contribution from each individual image should be recalibrated based on its relative importance to the others present in the set. In other words, each face image should be aware of the existence of the other input images. For instance, suppose all but one of the faces is frontal, then the non-frontal face could contribute more information relative to an individual frontal face. Thus the combination mechanism should also have the capability to compute the *relative* content quality of each face when determining its contribution to the final set-wise representation. These two quality measures are complementary: one is absolute, depending only on the individual image; the other is relative, depending on the set of images.

In this paper, we propose a Multicolumn Network (MN), that can be trained to aggregate the individual feature vectors into a set-wise descriptor and addresses both these deficiencies. The model is composed of a standard ResNet50 network with two extra quality control blocks, namely "visual" quality, and relative "content" quality, where the former block is able to down-weight the aberrant images, whilst the latter could highlight the most discriminative images when all images are of good "visual" quality. For efficient training, we first pre-train a ResNet50 on a large scale dataset–VGGFace2 [3] with standard image-wise classification loss, then use this ResNet as the backbone of the MN, and finetune the entire architecture end-to-end with set-wise classification. Overall, the proposed MN is shown to improve the performance on all of the IARPA IJB face recognition benchmarks over the previous state-of-the-art architectures.

## 2 Related Works

In this section, we review the work that has influenced the Multicolumn Network design.

**Attention-based feature aggregation.** Two recent papers have proposed architectures using multiple columns for face descriptor aggregation. They are Neural Aggregation Network (NAN) [19] and DR-GAN [17]. NAN uses within-set attention with a softmax for normalization, and aggregate from multiple inputs based on the softmax scores. The model also incorporates a second attention stage which inputs the aggregated feature descriptor (with the attention scores from the first stage) and outputs a filter kernel that is applied to each face descriptor in the set as a form of channel-wise modulation. DR-GAN generates a weight for each input using a (sigmoid) gating function, and aggregates from the multiple inputs using a weighted average. It does not have a second stage.

In a similar manner to these previous works, the proposed MN has a first stage that aggregates by weighted averaging (on visual "quality" for each input face image). In contrast, for the second stage, the MN models the relation between each input face and the aggregated feature, and produces different relative "content" quality scores for each face within the set (rather than a common modulation as in NAN). In this way the MN is designed to handle the challenges from both visual and content quality; and indeed, the proposed MN demonstrates superior performance over NAN and DR-GAN on the public benchmarks, as shown in the IJB-A results of section 4.3.

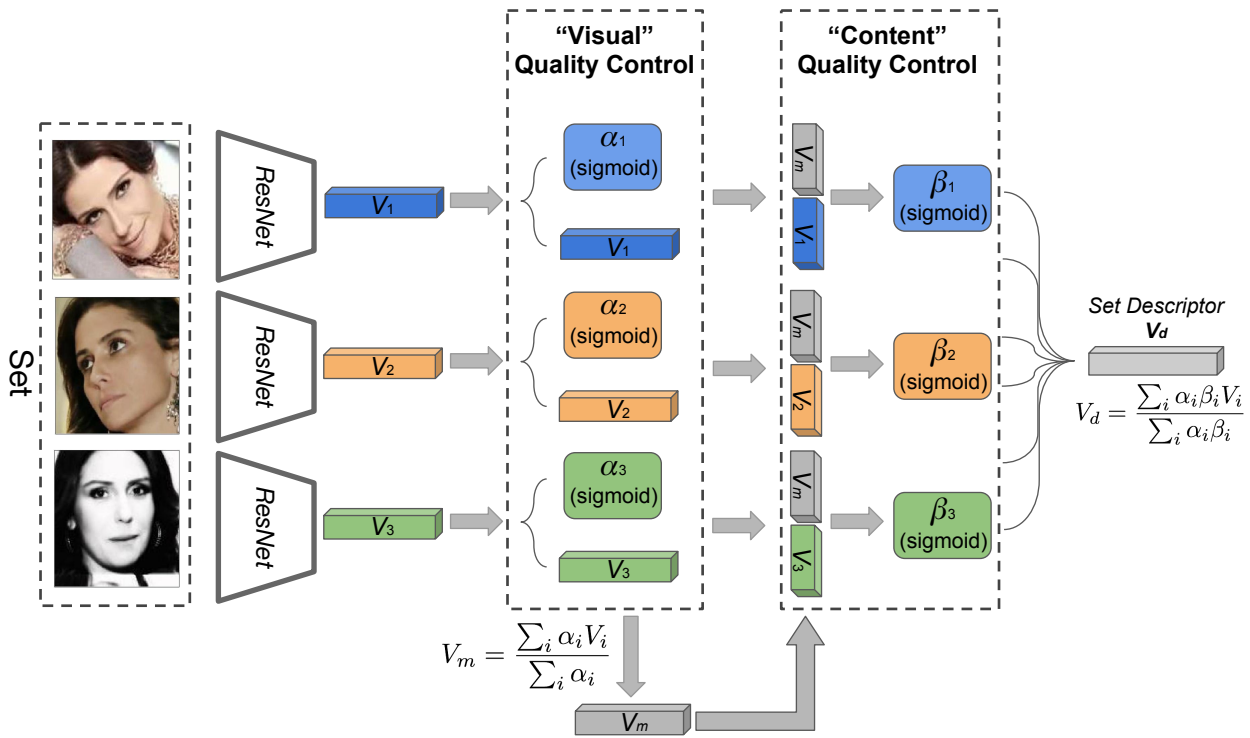

Figure 1: Overview of the Multicolumn Network (MN).
In order to obtain the feature descriptor for the entire set, all images are passed through an embedding module (parametrized as a ResNet50) until the last residual block. For "visual" quality control, scalar outputs ($\alpha$) are implicitly predicted with a fully connected (FC) layer for each image. For "content" quality control, the feature representation of each face ($V$) is concatenated with the mean face ($V_m$), and $\beta$ is predicted by a fully connected (FC) layer, therefore, the relative contribution of each face image is modelled explicitly. Eventually, the set descriptor is computed by incorporating both factors, i.e. "visual" quality and "content" quality. Note, although only three images are shown here, the architecture can ingest a variable number of images at test time.

**Relation/metric learning.** In [12], the authors propose a simple Relation Network (RN) for solving the problem of relational reasoning. In order to model the relations among objects, feature descriptors at every spatial location are concatenated with those at every other locations, and the RN module is parametrized with fully connected (fc) layers, which are explicitly trained to infer the existence and implications of object relations. In [15], the authors propose similar idea for low-shot learning. In order to model the relations (similarity) between images from a support set and test set, feature maps of the images are concatenated, and passed to a relation module (also parametrized as fc layers) for similarity learning. In the proposed Multicolumn Network, we take inspiration from both, and aim to model the relations between each image and every other image within the set.

## 3 Multicolumn Networks

We consider the task of set-based face verification, where the objective is to verify if two given sets are of the same person or not. We propose a Multicolumn Network (MN) that can ingest a variable number of images as input, and produce a fixed-size feature descriptor for the entire template. The descriptor is also permutation invariant to the order of the images. The MN architecture is shown in Figure 1. It consists of three sequential blocks:

*First*, a common embedding module (i.e. shared weights) is used to generate the per-image feature descriptor. This module is based on the standard ResNet50. Having a common module ensures that this stage of the architecture is permutation-invariant.

*Second*, the compact feature descriptor for each input image is passed to a quality assessment block, and a scalar inferred as a "visual" quality indicator. This ensures that the visual quality is computed independently for each each input image, and completes the task of the column per image. The outcome of the column is thus two-fold: the descriptor, and a weighting generated by a gating (sigmoid) function which can downweight aberrant images (e.g. low quality, extreme illumination, nonface).

*Third*, to compute the relative contribution of each image, the relations of each input image should ideally be modelled together with every other one within the set. However, this would introduce $\binom{n}{2}$ different combinations (*n* refers to the number of images within the set). To tackle this problem, we use a "mean face" as an anchor, and each face descriptor is only compared to this anchor to get their relative contribution to the final set descriptor.

In the following sections we give more implementation details, and the architecture is fully specified in Table 1. In the following, we say 'fully connected', but actually this is only a 1D weight vector (e.g. $2048 \times 1$) of the same size as the input, rather than the fc layers in AlexNet, which are huge (e.g. $4096 \times 4096$). Thus, the proposed MN actually only introduce a very small number of extra parameters, and can be used along with any popular architectures, e.g. VGGNets, ResNets.

## 3.1 Embedding Module

Formally, we denote the input image as $I_1, I_2, ..., I_n$, and the shared embedding is parameterized as a standard ResNet50 $\psi(\cdot, \theta_1)$, the outputs from the embedding module is therefore:

$$[V_1, V_2, ..., V_n] = [\psi(I_1; \theta_1), \psi(I_2; \theta_1), ..., \psi(I_n; \theta_1)] \qquad (1)$$

where each input image is of size $I_n \in R^{224 \times 224 \times 3}$, and the outputed compact feature representation $V_n \in R^{1 \times 2048}$.

## 3.2 Self-aware Visual Quality Assessment

In this part, we parametrize a quality control block with a simple fully conneted layer. Formally, we have:

$$[\alpha_1, \alpha_2, ..., \alpha_n] = [f(V_1; \theta_2), f(V_2; \theta_2), ..., f(V_n; \theta_2)] \qquad (2)$$

where $\theta_2 \in R^{2048 \times 1}$. In our case, we apply a sigmoid function: $f(\cdot, \theta_2) = \sigma(\cdot, \theta_2)$. Therefore, the block takes the compact feature vectors as inputs, and produce a scalar score ([0-1]) indicating low to high visual quality.

## 3.3 Content-aware Content Quality Assessment

To estimate the relative contribution of each face image, we first approximate the feature descriptor for the anchor face as :

$$V_m = \frac{\sum_{i=1}^{n} \alpha_i V_i}{\sum_{i}^{n} \alpha_i}, \quad (3)$$

the "content" quality is then estimated by modelling the relation between the individual inputs and the mean face, parametrized as $g(\cdot;\theta_3)$:

$$[\beta_1, \beta_2, ..., \beta_n] = [g([V_m{:}V_1];\theta_3), g([V_m{:}V_2];\theta_3), ..., g(V_m{:}V_n];\theta_3)] \quad (4)$$

where $[V_m{:}V_n]$ refers to the concatenation of the feature descriptor from the mean face and input image, $\theta_3 \in R^{4096 \times 1}$. Again, the function $g$ is parametrized with a fully connected layer, and a sigmoid function is used: $g(\cdot, \theta_3) = \sigma(\cdot, \theta_3)$.

## 3.4 Set-based Descriptor

As a result, the two extra quality control blocks introduce only about 6*K* more parameters to the standard ResNet50, and the set descriptor is then calculated as a combination of both "visual" and "content" qualities:

$$V_d = \frac{\sum_{i=1}^{n} \alpha_i \beta_i V_i}{\sum_{i}^{n} \alpha_i \beta_i}, \quad (5)$$

# 4 Experimental Details

## 4.1 VGGFace2 Dataset

In this paper, all models are trained with the large-scale VGGFace2 dataset [3]. In total, the dataset contains about 3.31 million images with large variations in pose, age, illumination, ethnicity and profession (e.g. actors, athletes, politicians). Approximately, 362.6 images exist for each of the 9131 identities on average. The entire dataset is divided into the training set (8631 identities) and validation set (500 identities). In order to be comparable with existing models, we follow the same dataset split and train the models with only 8631 identities.

## 4.2 Training Details

We train the Multicolumn Network (Table 1) in two phases:

*First*, we pretrain a ResNet50 with image-wise classification, at this step, a region of $224 \times 224$ pixels is randomly cropped from each sample, and the shorter side resized to 256. The mean value of each channel is subtracted for each pixel. Stochastic gradient descent is used with mini-batches of size 256, with a balancing-sampling strategy for each mini-batch due to the unbalanced training distributions. The initial learning rate is 0.1 for the models learned from scratch, and this is decreased twice with a factor of 10 when errors plateau. The weights of the models are initialised as described in [6]. This training process exactly follows the one described in [3].

*Second*, we further use the pre-trained networks in the Multicolumn Network, and finetune this model with set-based classification, each set contains 3 images (we also tried to train with 5 or 7 images as a set, however, the error signals become too weak for efficient training), random transformations are used with a probability of 20% for each image, e.g. monochrome augmentation, horizontal flipping, geometric transformation, gaussian blur, motion blur and jpeg compression.

| Module | ResNet-50 | Output size |
|---|---|---|
| Embedding | conv, $7 \times 7$, 64, stride 2 | $N \times 112 \times 112 \times 64$ |
| | max pool, $3 \times 3$, stride 2 | $N \times 56 \times 56 \times 64$ |
| | $\begin{bmatrix} \text{conv}, 1 \times 1, 64 \\ \text{conv}, 3 \times 3, 64 \\ \text{conv}, 1 \times 1, 256 \end{bmatrix} \times 3$ | $N \times 56 \times 56 \times 256$ |
| | $\begin{bmatrix} \text{conv}, 1 \times 1, 128 \\ \text{conv}, 3 \times 3, 128 \\ \text{conv}, 1 \times 1, 512 \end{bmatrix} \times 4$ | $N \times 28 \times 28 \times 5125$ |
| | $\begin{bmatrix} \text{conv}, 1 \times 1, 256 \\ \text{conv}, 3 \times 3, 256 \\ \text{conv}, 1 \times 1, 1024 \end{bmatrix} \times 6$ | $N \times 14 \times 14 \times 1024$ |
| | $\begin{bmatrix} \text{conv}, 1 \times 1, 512 \\ \text{conv}, 3 \times 3, 512 \\ \text{conv}, 1 \times 1, 2048 \end{bmatrix} \times 3$ | $N \times 7 \times 7 \times 2048$ |
| | global average pool | $N \times 1 \times 1 \times 2048$ $(V_1...V_N)$ |
| Visual Quality | fc, $2048 \times 1$ | $N \times 1$ $(\alpha_1...\alpha_N)$ |
| | weighted average | $1 \times 1 \times 1 \times 2048$ $(V_m = \frac{\sum_{i=1}^{n} \alpha_i V_i}{\sum_i^n \alpha_i})$ |
| Content Quality | feature concatenation | $N \times 1 \times 1 \times 4096$ (V $= [V_N : V_m]$) |
| | fc, $4096 \times 1$ | $N \times 1$ $(\beta_1...\beta_N)$ |
| | weighted average | $1 \times 1 \times 1 \times 2048$ $(V_d = \frac{\sum_{i=1}^{n} \alpha_i \beta_i V_i}{\sum_i^n \alpha_i \beta_i})$ |

Table 1: Multicolumn Networks (MN).
In order to obtain the feature descriptor for the N input images (N=3 in our case), all images are passed through an embedding module (parametrized as a ResNet50) until the last residual block. For "visual" quality control, scalar outputs ($\alpha$) are implicitly predicted with a fully connected (FC) layer for each image. For "content" quality control, the feature representation of each face ($V$) is concatenated with the mean face ($V_m$), and $\beta$ is predicted by a fully connected (FC) layer, enabling the relative contribution of each face image to be modelled explicitly. Eventually, the set descriptor is computed by incorporating both factors, i.e. "visual" quality and "content" quality.

## 4.3 Results

We evaluate all models on the challenging IARPA Janus Benchmarks, where all images and videos are captured from unconstrained environments with large variations in viewpoints and image quality. We evaluate the models on the standard 1:1 verification protocol, where a set consists of a variable number of face images and video frames from different sources. (i.e. each set can be image-only, video-frame-only, or a mixture of still images and frames). Note, in contrast to the traditional closed-world classification tasks (where the identities are the same during training and testing), face verification is treated as an open-world problem (i.e. the label spaces of the training and test set are disjoint), and thus challenges the capacity and generalization of the feature representations.

During testing, set descriptor is computed and compared with cosine similarity. The performance is reported as the true accept rates (TAR) vs. false positive rates (FAR) (i.e. receiver operating characteristics (ROC) curve). To evaluate the effect of different quality assessment blocks, two versions of the architecture are compared, namely MN-v ("visual" quality only) and MN-vc ("visual" + "content"), as described in equations (3) and (5).

**IJB-A Dataset [8]** The IJB-A dataset contains 5712 images and 2085 videos from 500 subjects, with an average of 11.4 images and 4.2 videos per subject.

|  | Architecture | Dataset | 1:1 Verification TAR | | |
| --- | --- | --- | --- | --- | --- |
|  |  |  | FAR=$1E-3$ | FAR=$1E-2$ | FAR=$1E-1$ |
| DR-GAN [17] | CASIA-Net [20] | Multi-PIE+CASIA-WebFace | $0.539 \pm 0.043$ | $0.774 \pm 0.027$ | – |
| Bansalit et al. [1] | VGGNet | UMDFace | $0.770^\dagger$ | 0.893 | $0.960^\dagger$ |
| Crosswhite et al. [4] | VGGNet | VGGFace | $0.836 \pm 0.027$ | $0.939 \pm 0.013$ | $0.979 \pm 0.004$ |
| NAN ([19]) | GoogLeNet | Indoor Data (3M Images) | $0.881 \pm 0.011$ | $0.941 \pm 0.008$ | $0.978 \pm 0.003$ |
| Cao et al. [3] | ResNet50 | VGGFace2 | $0.895 \pm 0.019$ | $0.950 \pm 0.005$ | $0.980 \pm 0.003$ |
| Cao et al. [3] | SENet50 | VGGFace2 | $0.904 \pm 0.020$ | $0.958 \pm 0.004$ | $0.985 \pm 0.002$ |
| MN-v | ResNet50 | VGGFace2 | $0.906 \pm 0.020$ | $0.959 \pm 0.004$ | $0.987 \pm 0.002$ |
| MN-vc | ResNet50 | VGGFace2 | **$0.920 \pm 0.013$** | **$0.962 \pm 0.005$** | **$0.989 \pm 0.002$** |

Table 2: Evaluation on 1:1 verification protocol on IJB-A dataset. Higher is better.
Results from Cao et al. are computed by the commonly used feature aggregation: averaging.
The values with † are read from the ROC curve in [1].

**IJB-B Dataset [18]** The IJB-B dataset is an extension of IJB-A [8], having $1,845$ subjects with 21.8K still images (including $11,754$ face and $10,044$ non-face) and 55K frames from $7,011$ videos.

|  | Architecture | Dataset | 1:1 Verification TAR | | | | |
| --- | --- | --- | --- | --- | --- | --- | --- |
|  |  |  | FAR=$1E-5$ | FAR=$1E-4$ | FAR=$1E-3$ | FAR=$1E-2$ | FAR=$1E-1$ |
| Whitelam et al. [18] | VGGNet | VGGFace | 0.380 | 0.540 | 0.700 | 0.840 | 0.950 |
| Navaneeth et al. [2] | – | UMDFace | – | 0.685 | 0.830 | 0.925 | 0.978 |
| Cao et al. [3] | ResNet50 | VGGFace2 | 0.647 | 0.784 | 0.878 | 0.938 | 0.975 |
| Cao et al. [3] | SENet50 | VGGFace2 | 0.671 | 0.800 | 0.888 | 0.949 | 0.984 |
| MN-v | ResNet50 | VGGFace2 | 0.683 | 0.818 | 0.902 | 0.955 | 0.984 |
| MN-vc | ResNet50 | VGGFace2 | **0.708** | **0.831** | **0.909** | **0.958** | **0.985** |

Table 3: Evaluation on 1:1 verification protocol on IJB-B dataset. Higher is better.
Note that the result of Navaneeth et al. [2] is on the Janus CS3 dataset.

**IJB-C Dataset [10]** The IJB-C dataset is a further extension of IJB-B, having $3,531$ subjects with 31.3K still images and 117.5K frames from $11,779$ videos. In total, there are 23124 templates with 19557 genuine matches and $15639K$ impostor matches.

## 4.4 Discussion

The IJB-A benchmark only contains a very small number of testing faces (about $10K$ pairs), and this benchmark has already become saturated. In this section, we will mainly focus on discussing the results from the more challenging IJB-B and IJB-C benchmarks. The following phenomena can be observed from the evaluation results:

*First*, comparing with the baseline ResNet50, the proposed MN architecture (using the same backbone ResNet50) only introduces an extra $6K$ parameters, but the performance is consistently boosted by about 2-6% on both IJB-B and IJB-C.

|  | Architecture | Dataset | 1:1 Verification TAR ||||| 
|---|---|---|---|---|---|---|---|
|  |  |  | FAR=$1E-5$ | FAR=$1E-4$ | FAR=$1E-3$ | FAR=$1E-2$ | FAR=$1E-1$ |
| GOTS-1 [10][†] | – | – | 0.090 | 0.160 | 0.320 | 0.620 | 0.800 |
| FaceNet [10][†] | Inception | Indoor Data | 0.330 | 0.490 | 0.660 | 0.820 | 0.920 |
| VGG-CNN [10][†] | VGGNet | VGGFace | 0.430 | 0.600 | 0.750 | 0.860 | 0.950 |
| Cao *et al.* [3] | ResNet50 | VGGFace2 | 0.734 | 0.825 | 0.900 | 0.950 | 0.980 |
| Cao *et al.* [3] | SENet50 | VGGFace2 | 0.747 | 0.840 | 0.910 | 0.960 | 0.987 |
| MN-v | ResNet50 | VGGFace2 | 0.755 | 0.852 | 0.920 | 0.965 | 0.988 |
| MN-vc | ResNet50 | VGGFace2 | **0.771** | **0.862** | **0.927** | **0.968** | **0.989** |

Table 4: Evaluation on 1:1 verification protocol on IJB-C dataset. Higher is better. The results marked with † are read from the ROC curve in [10].

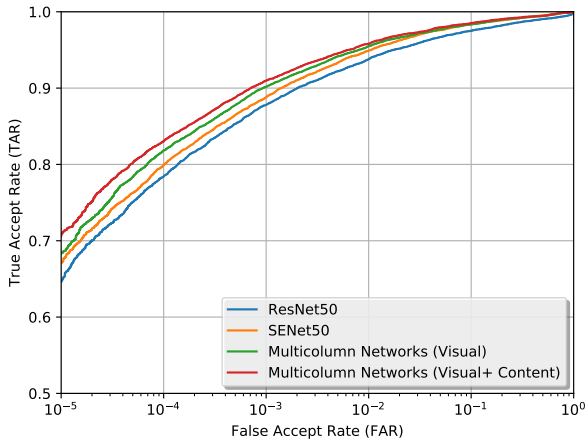
(a) ROC for IJB-B

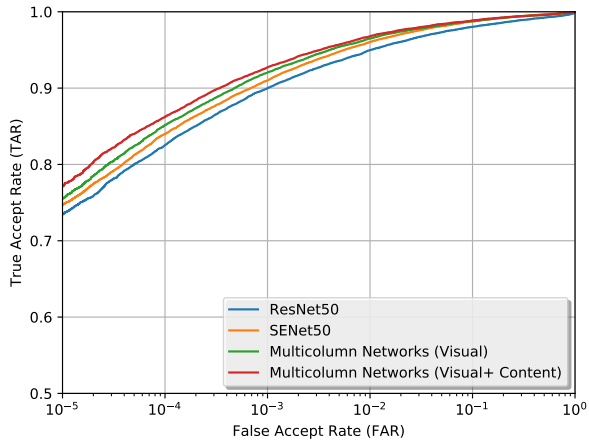
(b) ROC for IJB-C

Figure 2: ROC curve of 1:1 verification protocol on IJB-B & IJB-C dataset.
Comparing with the baseline (ResNet50) and the state-of-the-art model (SENet-50), the Multicolumn Networks consistently show an improvement on the benchmarks.

*Second*, MN-v (visual only) already beats the state-of-the-art architecture (SENet50), showing the importance of suppressing the aberrant images within the set. This architecture is similar to that proposed in DR-GAN [17]. However, as shown in the IJB-A results (Table 2), the MN-v results are significantly better than those of DR-GAN; this shows the benefit of using a stronger backbone architecture (i.e. ResNet50 compared to CASIA-Net).

*Third*, when adding the "content" assessment (MN-vc), the performance is further improved. This indicates that both "visual" and "content" assessments actually play a role in boosting the verification results.

*Fourth*, performance improvements are most substantial at $FAR = 10^{-5}$, $FAR = 10^{-4}$ and $FAR = 10^{-3}$ (left starting part of both ROC curves in Figure 2). This is as expected due to the fact that the main issue of average feature aggregation (as used in previous results) is that the set-based descriptor can be distracted by low-quality images (either "visual" or "content"). The consequence is that they can then match to any other set, e.g. blurry images of different people can often look very similar, extreme pose faces or covered faces can look similar. Consequently, the matching scores for the false positives pairs can be higher than those of true positives. By suppressing the aberrant images and highlighting the discriminative faces, the most dramatic improvement that MN-vc brings to producing high quality set descriptors is in reducing the false positive rate.

Moreover, we would also expect to see a further performance boost by using a more powerful backbone architecture. As a proof of concept, we take the publicly available SENets_ft from Cao *et al.* [3] (pretrained on both MS1M [5] and VGGFace2), extract features for the images in the IJB-C benchmark, and simply use the "visual" quality scores from the Multicolumn Network (trained with ResNet50) to generate the set descriptors. As shown in Table 5, compared with the naïve feature aggregation this already shows a significant improvement of over 5% and 2% at $FAR = 10^{-5}$ and $FAR = 10^{-4}$, respectively, on the very strong baseline. This supports once again the false positive discussion above, and shows the potential benefit of using a stronger backbone network for future end-to-end training of the Multicolumn Network.

|  | Architecture | Dataset | 1:1 Verification TAR | | | |
| --- | --- | --- | --- | --- | --- | --- |
|  |  |  | FAR=$1E-5$ | FAR=$1E-4$ | FAR=$1E-3$ | FAR=$1E-2$ |
| Cao *et al.* [3] | SENet50 | MS1M+VGGFace2 | 0.768 | 0.862 | 0.927 | 0.967 |
| MN-v | SENet50 | VGGFace2 | **0.819** | **0.887** | **0.938** | **0.968** |

Table 5: Evaluation on 1:1 verification protocol on IJB-C dataset. Higher is better.

## 4.5 Visualization

In Figure 3, we first show the sorted images in ascending order based on the scores implicitly inferred from the self-aware "visual" quality assessment block. As expected, the self-aware "visual" quality scores for aberrant images are highly correlated with human intuition, i.e. blurry, nonface, extreme poses. Note, the images of medium and high quality are not so well separated, though high quality ones are often near frontal.

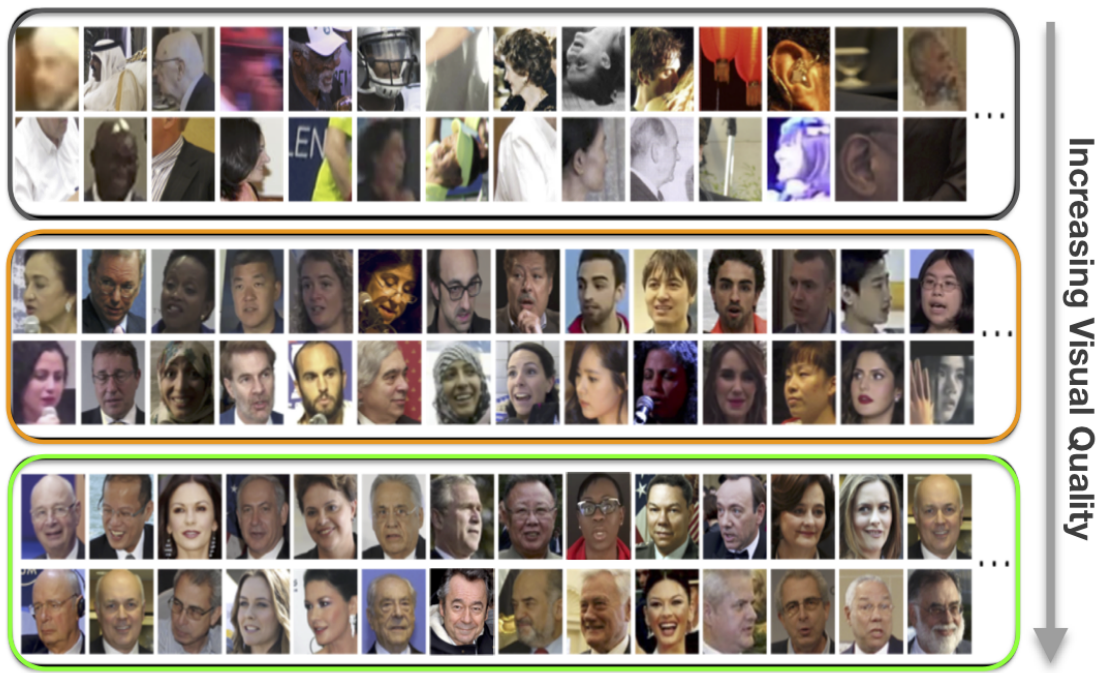

Figure 3: Visualization of the "visual" quality for the images in the IJB benchmark.
Following the arrow, the sigmoid scores ($\alpha$) get higher. From the perspective of the Multicolumn Network, those bottom images are treated as of higher "visual" quality than the top images. As expected, this is highly correlated with the way we define good face images.

In Figure 4, we show the images and "content" scores ($\gamma$) based on the content-aware quality assessment block. Two observations can be drawn to show the benefit of having double quality assessment blocks in MNs: *First*, images with relatively lower "visual" quality may possess more importance than the higher ones, e.g. image sets in the first row. *Second*, images with similar "visual" quality can be of very different importances, e.g. image sets in the 3rd row.

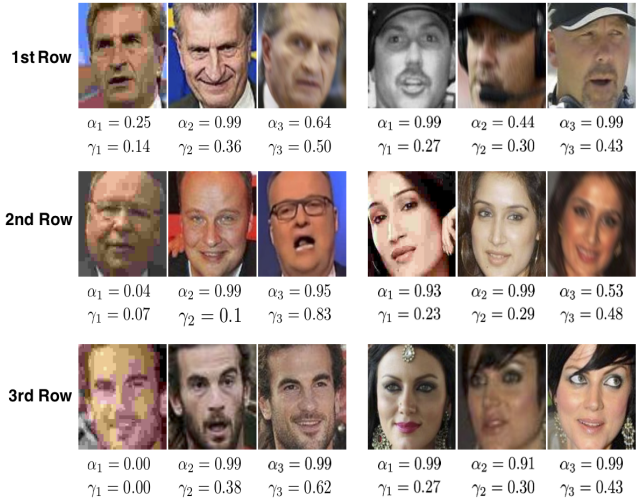

Figure 4: Visualization of the image sets after within-set recalibration. We use three images (from VGGFace2) as a set, and $\alpha$ and $\gamma$ refer to the "visual" quality and the recalibrated importance, i.e. $\gamma_i = \frac{\alpha_i \beta_i}{\sum_j \alpha_j \beta_j}$. The images within the set are ordered based on $\gamma$. As can be seen, the images with relatively higher "visual" quality do not necessarily indicate higher importance while calculating the final set-wise feature, e.g. images sets in the first row. Images with similar "visual" quality can be of very different importances, e.g. image sets in the 3rd row.

## 5 Conclusions

We have demonstrated that the addition of two intuitive weightings (visual and content) can significantly improve the verification performance of set based descriptors, compared to simple averaging. This addition only adds around 6K more parameters to the 25.6M parameters of the ResNet50 architecture. As future works, there are several promising directions: *First*, a more powerful backbone network can be used for end-to-end training of Multicolumn Networks, which can further boost the performance. *Second*, to further improve the performance, quality assessment should not only be based on the feature descriptor from the whole face, but also be focusing on more detailed facial parts, e.g. eyes, noses. Therefore, for each input set, several feature descriptors can be computed, weighted by visual and content qualities, and trained with set-wise classification.

## Acknowledgment

We benefited from insightful discussions with Lars Ericson, Jeffrey Byrne, Chris Boehnen, Patrick Grother. This research is based upon work supported by the Office of the Director of National Intelligence (ODNI), Intelligence Advanced Research Projects Activity (IARPA), via contract number 2014-14071600010. The views and conclusions contained herein are those of the authors and should not be interpreted as necessarily representing the official policies or endorsements, either expressed or implied, of ODNI, IARPA, or the U.S. Government. The U.S. Government is authorized to reproduce and distribute reprints for Governmental purpose notwithstanding any copyright annotation thereon.